\documentclass[journal]{IEEEtran}
\usepackage[cmex10]{amsmath}
\usepackage{amsfonts}
\usepackage{graphicx}
\usepackage{array}
\usepackage{multirow}
\usepackage{url}
\usepackage{epstopdf}
\usepackage{color}
\usepackage[switch]{lineno}

\usepackage{bm}
\usepackage{color}

\begin{document}
\title{RGBT Salient Object Detection: A Large-scale Dataset and Benchmark
}

\author{Zhengzheng~Tu, Yan~Ma, Zhun~Li, Chenglong~Li, Jieming~Xu, Yongtao~Liu
 \thanks{Z. Tu, Y. Ma, Z. Li, C. Li, J. Xu, and Y. Liu are with Anhui Provincial Key Laboratory of Multimodal Cognitive Computation, School of Computer Science and Technology, Anhui University, Hefei, China, Email: zhengzhengahu@163.com, m17856174397@163.com, 18355607109@163.com, lcl1314@foxmail.com, m17730263582@163.com, lyt494131144@163.com. C. Li is also with Institute of Physical Science and Information Technology, Anhui University, Hefei 230601, China. (\emph{Corresponding author is Chenglong Li})}
\thanks{This research is jointly supported by Natural Science Foundation of Anhui Higher Education Institution of China(No.KJ2020A0033), Anhui Provincial Natural Science foundation(No.2108085MF211,2020 Anhui Energy Internet Joint Fund Project: No.2008085UD07),National Natural Science Foundation of China (No.61876002), Anhui Provincial Key Research and Development Program (No.202104d07020008), NSFC Key Project of International (Regional) Cooperation and Exchanges (No. 61860206004).
}}

\markboth{IEEE Transactions on Multimedia}%
{Shell \MakeLowercase{\textit{et al.}}: Bare Demo of IEEEtran.cls for IEEE Journals}
\maketitle

\begin{abstract}
Salient object detection in complex scenes and environments is a challenging research topic.
Most works focus on RGB-based salient object detection, which limits its performance of real-life applications when confronted with adverse conditions such as dark environments and complex backgrounds.
Taking advantage of RGB and thermal infrared(RGBT) images becomes a new research direction for detecting salient objects in complex scenes, since the thermal infrared spectrum provides the complementary information and has been used in many computer vision tasks. 
However, current research for RGBT salient object detection is limited by the lack of a large-scale dataset and comprehensive benchmark.
This work contributes such a RGBT image dataset named VT5000, including 5000 spatially aligned RGBT image pairs with ground truth annotations.
VT5000 has 11 challenges collected in different scenes and environments for exploring the robustness of algorithms.
With this dataset, we propose a powerful baseline approach, which extracts multi-level features of each modality and aggregates these features of all modalities with the attention mechanism for accurate RGBT salient object detection. To further solve the problem of blur boundaries of salient objects, we also use an edge loss to refine the boundaries.
Extensive experiments show that the proposed baseline approach outperforms the state-of-the-art methods on VT5000 dataset and other two public datasets.
In addition, we carry out a comprehensive analysis of different algorithms of RGBT salient object detection on VT5000 dataset, and then make several valuable conclusions and provide some potential research directions for RGBT salient object detection. Our new VT5000 dataset is made publicly available at \textcolor{magenta}{https://github.com/lz118/RGBT-Salient-Object-Detection}.
\end{abstract}

\begin{IEEEkeywords}
Salient object detection, Attention, VT5000 dataset.
\end{IEEEkeywords}

\section{Introduction}
\label{sec::introduction}
\IEEEPARstart{S}{alient} object detection aims to find the object that human eyes pay much attention to in an image.
Salient object detection has been extensively studied over the past decade, but still faces many challenges in complex environment, e.g., when appearance of the object is similar to the surrounding, the algorithms of salient object detection for RGB images often perform not well.
Researches on adopting different modalities to assist salient object detection have attracted more and more attentions.
Many works~\cite{CiptadiHR13,JuGGRW14} have achieved good results on salient object detection by combining RGB images with depth information.
However, depth image has its limitations, for example, when the object is perpendicular to the shot of depth camera, the depth map shows inconsistent values of the same surface.
Integrating RGB and thermal infrared (RGBT) data has also shown its effectiveness for some computer vision tasks, such as moving object detection, person Re-ID, and visual tracking ~\cite{LiCHLTL16, abs-1805-08982, Wang0MZTL18}.
The imaging principle of thermal infrared camera is based on thermal radiation from the object surface, and different places of object surface have almost same thermal radiation.
Existing RGB and Depth(RGBD) salient object detection methods mainly focus on how to explore depth information to complement RGB SOD. However, RGBT salient object detection treats RGB and thermal modalities equally and its target is to leverage complementary advantages to discover common conspicuous objects in both modalities.

Recently, RGBT salient object detection has become attractive.
The first work of RGBT salient object detection~\cite{Wang0MZTL18} proposes a multi-task manifold ranking algorithm for RGBT image saliency detection, and creates an unified RGBT dataset called VT821. In addition, VT821 has several limitations:  (1) RGB and thermal imaging parameters are completely different, and there might thus be some alignment errors;
(2) The operators for modalities alignment will introduce the black background in image boundaries;
(3) Most of these scenarios are very simple, and thus it is not so challenging.
The second important work of RGBT image saliency detection~\cite{abs-1905-06741} contributes a more challenging dataset named VT1000 and proposes a novel collaborative graph learning algorithm.
Compared with VT821, VT1000 has its advantages but also have several limitations: (1) As RGB and thermal infrared imaging have different sighting distances, thermal infrared image and visible light image look different and need to be aligned, as shown as the left image pair in Fig.~\ref{camera};
(2) RGB image and thermal infrared image are still not automatically aligned, inevitably introducing errors in the process of manually aligning them;
(3) Although VT1000 is larger than VT821, the complexity and diversity of the scenes have not been greatly improved.
In this paper, we construct a more comprehensive benchmark for RGBT salient object detection based on the demands of large-scale, good resolution, high diversity and low bias.
First, existing RGBT datasets are not enough to train a good deep network, thus we collect 5000 pairs of RGB and thermal images in different environments, in which each pair of RGBT images is automatically aligned. 
Second, as most of the backgrounds or scenes are simple in existing datasets, our dataset considers different sizes, categories, surroundings, imaging quantities and spatial locations of salient objects, and we also give a statistical analysis to show the diversity of objects.
To analyze the sensitivity of different methods for various challenges, we annotate 11 different challenges in the consideration of above factors.
Finally, we annotate not only attributes of challenges but also imaging quality of objects in the dataset. The annotations of imaging quality of objects provide the labels for weakly supervised RGBT salient object detection in the future researches.

Although significant progress has been made in terms of RGBT saliency detection~\cite{Wang0MZTL18,abs-1905-06741,tu2019m3s}, the performance might be limited by three main problems: (1) Both of these works are based on traditional methods, but they simply merge the features of the two modalities without proposing an effective way for feature fusion ; (2) Background noise is introduced when the features of the two modalities are fused; (3) The problem of unclear object boundaries is not solved effectively.
To provide a powerful baseline for RGBT salient object detection, we design an end-to-end trained CNN-based framework.
In specific, the two-stream CNN architecture employs VGG16~\cite{simonyan2014very} as the backbone network to extract multi-scale RGB and thermal infrared features separately.
To obtain task-related features, we use channel-wise and spatial-wise attention based Convolution Block Attention Module (CBAM)~\cite{WooPLK18} to selectively collect features from RGB and thermal infrared branches.
Then we perform a pixel-wise addition on RGB and thermal infrared features to fuse them, and pass the merged features from first convolutional block of VGG16 to next convolutional block.
To obtain global guidance information, we input the fused RGB and thermal infrared features from last convolutional block into the Pyramid Pooling Module (PPM)~\cite{abs-1904-09569}.
We also use the average pooling to capture the global context information, thus can locate salient objects accurately.
To make better use of characteristics of different layers, we upsample the features processed by each block of VGG16 with different sampling rates, and then combine them with the features processed by PPM.
We also utilize the Feature Aggregation Module (FAM)~\cite{abs-1904-09569} after feature fusion to capture the local context information.
Finally, to solve the problem of unclear object boundaries, we employ an edge loss to guide the network to learn more details around boundaries, and further refine the boundary of the object.

In summary, the main contributions of this work are summarized as follows:

\begin{itemize}
	\item We create a large-scale RGBT dataset containing 5000 pairs of RGB and thermal images for salient object detection, with manually labeled ground truth annotations.
	We hope that this dataset would promote the research progress for deep learning techniques on RGBT salient object detection.
	This dataset and its all annotations will be released to public for free academic usage. 
	
	\item We propose a novel deep CNN architecture to provide a powerful baseline approach for RGBT salient object detection.
	In particular, we propose to utilize a convolutional block attention to selectively collect features from RGB and thermal infrared branches and an edge loss to guide the network to learn more details that can reserve the boundaries of salient objects.

	\item Extensive experiments show that the designed approach outperforms the state-of-the-art methods on VT5000 dataset and other two public datasets, i.e., VT821 and VT1000.
	In addition, a comprehensive analysis for different algorithms of RGBT salient object detection is performed on VT5000 dataset.
	Through the analysis, we make several valuable conclusions and provide some potential research directions for RGBT salient object detection.
\end{itemize}

\section{Related Work}
\subsection{Multi-modal Salient Object Detection Datasets}
\label{sec::2A}
With the emergence of multi-modal data, RGBD salient object detection (SOD) has been proposed, and the related RGBD datasets have been constructed.

NJU2K~\cite{ju2014depth} collects 1,985 RGBD image pairs which are from the Internet and 3D movies or taken by a Fuji W3 stereo camera. 
NLPR~\cite{peng2014rgbd} uses Microsoft Kinect captures 1,000 RGBD image pairs.
DES~\cite{cheng2014depth} also uses Microsoft Kinect to collect 135 RGBD image pairs in indoor scenes, and it is also called RGBD135.
SSD~\cite{zhu2017three} contains 80 image pairs picked up from three stereo movies.
STERE~\cite{niu2012leveraging} has 1000 image pairs collected in real-world scenes and virtual scenes.
More specifically, GIT~\cite{CiptadiHR13} and LFSD~\cite{LiYJLY17} datasets are designed for the specific purposes, such as generic object segmentation based on saliency map or saliency detection in the light field.

Subsequently, Li et al.~\cite{Wang0MZTL18} construct the first RGBT dataset VT821 with 821 pairs of RGBT images.
Tu et al.~\cite{abs-1905-06741} contribute a more challenging dataset VT1000 for RGBT image saliency detection.
\subsection{Attention Mechanism}
Attention mechanism is first proposed by Bahdanau et al.~\cite{BahdanauCB14} for neural machine translation, the attention mechanisms in deep neural networks have been studied widely recently. Attention mechanisms are proven to be useful in many tasks, such as scene recognition~\cite{CaoLYYWWHWHXRH15,HongYKH15}, question answering~\cite{YangHGDS16}, caption generation~\cite{XuBKCCSZB15} and pose estimation~\cite{ChuYOMYW17}. For example, Chu et al.~\cite{ChuYOMYW17} propose a network based on multi-context attention mechanism and apply it to the end-to-end framework of pose estimation. Zhang et al.~\cite{ZhangWQLW18} propose a progressive attention guidance network, which generates attention features successively through the channel and the spatial attention mechanisms for salient object detection. In PiCANet~\cite{LiuH018}, Liu et al. propose a novel pixel-wise contextual attention network. In specific, the network generates the attention map with the contextual information of each pixel. With the learned attention map, the network selectively incorporates the features of useful contextual locations, thus contextual features can be constructed. Then the pixel-wise contextual attention mechanism is embedded into the pooling and convolution layers to bring in the global or local contextual information.

As performing quite well on feature selection, the attention mechanism is also suitable for salient object detection. Some methods adopt effective strategies, such as progressive attention~\cite{ZhangWQLW18} and gate function~\cite{ZhangDLH018}. Inspired by the above, we utilize a lightweight and general attention module~\cite{WooPLK18}, which decomposes the learning process of channel-wise attention and spatial-wise attention. The separate attention generation process of the 3D feature map has much less parameters and computational cost.

Moreover, in order to enhance the ability of feature representation, then channel-wise attention mechanism assigns the weight to the channel that is highly responsive to the salient object. Some details in the background are inevitably introduced when the saliency map is generated with low-level features. Taking advantages of high-level features, spatial-wise attention mechanism removes some backgrounds, thus highlights foreground area, which benefits salient object detection.

\subsection{Multi-modal SOD methods}
In recent years, with the popularity of thermal sensors, integrating RGB and thermal infrared data has applied to many tasks of computer vision ~\cite{LiZLZT17, LiZHTW18, LiuS12, Wang0MZTL18, YangLLWT18}.

In addition to RGBT SOD, there are many methods adopting different modality to obtain multiple cues for better detection, such as depth images. 
In order to combine multiple modalities well, many RGBD methods utilize the better modality to assist the other modality. 
For example, Qu et al.~\cite{QuHZTTY17} design a novel network to automatically learn the interaction mechanism for RGBD salient object detection.
Han et al.~\cite{HanCLYL18} design a two-stream architecture, combining the depth representation to make the collaborative decision through a joint full connection layer.
Wang et al.~\cite{wang2019adaptive} also propose a two-stream network, in which they design a saliency fusion module to learn a switch map for  fusing the saliency maps adaptively.
Chen et al.~\cite{chen2019multi} realize cross-modal interactions by designing a multi-scale multi-path fusion network. They diversify the fusion path to perform the global reasoning and the local capturing. Compared with two-stream architectures, this method has more adaptive and flexible fusion flows.
Chen et al.~\cite{chen2018progressively} also propose to learn complementary information by designing a novel complementarity-aware fusion module with cross-modal residual functions and complementarity-aware supervisions. They cascade this module and add the level-wise supervision, which enables more sufficient fusion results.
Piao et al.\cite{piao2019depth} design a depth-induced multi-scale recurrent attention network. They use residual connections to extract and fuse multi-level paired complementary cues. Then they combine depth cues with multi-scale context features for accurate location. Finally, they use a recurrent attention module to generate more accurate results.
Piao et al.\cite{piao2020a2dele} further consider that existing two-stream methods have extra costs, bringing difficulties to practical applications. So they propose a depth distiller to transfer the depth knowledge from the depth stream to the RGB stream, thus construct a lightweight architecture that is practically usable.
Inspired by the non-local model, Liu et al.\cite{liu2020learning} propose to integrate the self-attention and the mutual attention to propagate long range contextual dependencies,thus incorporate multi-modal information for more accurate results. 
Compared with the thermal infrared camera, depth imaging has the limitation that the objects with the same distance to the camera have the same gray level, which is an obvious weakness of depth images. In addition, RGB imaging is usually influenced by various illuminations or bad weathers.
To avoid the above problems, more and more researches focus on how to fuse RGB and thermal infrared images, and take full advantages of complementary advantages to discover common conspicuous objects in both modalities.
For example, Wang et al.~\cite{Wang0MZTL18} propose a multi-task manifold ranking algorithm for RGBT image saliency detection, and at the same time build up an unified RGBT image dataset.
Tu et al.~\cite{abs-1905-06741} propose an effective RGBT saliency detection method by taking superpixels as graph nodes, moreover, they use hierarchical deep features to learn the graph affinity and the node saliency in a unified optimization framework.
With this benchmark~\cite{Wang0MZTL18}, Tang et al.~\cite{tang2019rgbt} propose a novel approach based on a cooperative ranking algorithm for RGBT salient object detection, they introduce a weight for each modality to describe the reliability and design an efficient solver for multiple subproblems.
All of the above methods are based on traditional methods, and time-consuming. Therefore, we propose an end-to-end deep network for RGBT salient object detection.

\section{VT5000 Benchmark}
In this work, for promoting the research of RGBT salient object detection(SOD) and considering the insufficiency of existing RGBT datasets, we capture 5000  pairs of RGBT images. In this section, we will introduce our new dataset in detail.

\subsection{Capture Platform}
The equipment used to collect RGB and thermal infrared images are FLIR (Forward Looking Infrared) T640 and T610, as shown in Fig.~\ref{camera}, equipped with the thermal infrared camera and CCD camera.
The two cameras have same imaging parameters, thus we don't need to manually align RGB and thermal infrared images one by one, which reduces errors from manual alignment.

\subsection{Data Annotation}
In order to evaluate RGBT SOD algorithms comprehensively, after collecting more than 5500 pairs of RGB images and corresponding thermal infrared images, we first select 5500 pairs of RGBT images as different as possible, each of image pairs contains one or more salient objects.
Similar to many popular SOD datasets~\cite{AchantaHES09,ChengMHTH15,JiangCLBW19,LiXLY17,XiaLCZZ17}, we ask six viewers to choose the most salient objects they saw at the first sight for the same image.
Because different person might look at different salient object in the same image, 5000 pairs of RGBT images with same selection for salient objects are finally retained.
Finally, we use Adobe Photoshop to manually segment the salient objects in each image to obtain pixel-level ground truth masks.

\begin{table*}[!ht]\normalsize
	\caption{Distribution of attributes and imaging quality in VT5000 dataset, showing the number of coincident attributes across all RGBT image pairs. The last two rows and two columns in the Table.\ref{attributes} indicate the poor performance of RGB and T respectively,due to low light, out of focus and thermal crossover, etc.}\label{attributes}
	\begin{center}
		\setlength{\abovecaptionskip}{5pt}
		\setlength{\belowcaptionskip}{5pt}
		\begin{tabular}{c|cccccccccccccc}
			\hline
			{CHALLENGE} &{BSO} &{CB} &{CIB} &{IC}&{LI}&{MSO}&{OF}&{SSO}&{SA}&{TC}&{BW}&{RGB}&{T}\\
			\hline
			{BSO} &\textbf{1746}  &371 &590&446  &211  &159 &96 &3 &99 &244&66&138&206\\
			{CB} &371  &\textbf{1176} &388&286&113&224&62&88&65&177&45&78&151\\
			{CIB} &590 &388  &\textbf{1134}&292&112&114&49&13&70&148&76&70&123\\
			{IC} &446  &286 &292&\textbf{1096}&66&80&54&38&46&193&57&62&160\\
			{LI} &211 &113 &112&66&\textbf{535}&51&89&22&71&83&23&188&70\\
			{MSO} &159 &224 &114&80&51&\textbf{491}&35&43&48&83&27&42&69\\
			{OF} &96 & 62 &49&54&89&35&\textbf{326}&21&18&73&20&177&65\\
			{SSO} &3 &88 &13&38&22&43&21&\textbf{340}&37&77&6&20&66\\
			{SA} &99 &65 &70&46&71&48&18&37&\textbf{401}&76&16&74&64\\
			{TC} &244 & 177 &148&193&83&83&73&77&76&\textbf{858}&32&90&629\\
			{BW} &66 & 45 &76&57&23&27&20&6&16&32&\textbf{222}&28&40\\
			{RGB} &138 & 78&70&62&188&42&177&20&74&90&28&\textbf{401}&92\\
			{T} &206 &151&123&160&70&69&65&66&64&629&40&92&\textbf{702}\\
			\hline
		\end{tabular}
	\end{center}
\end{table*}

\begin{figure*}[htbp]
	\centering
	\includegraphics[width=7.0in]{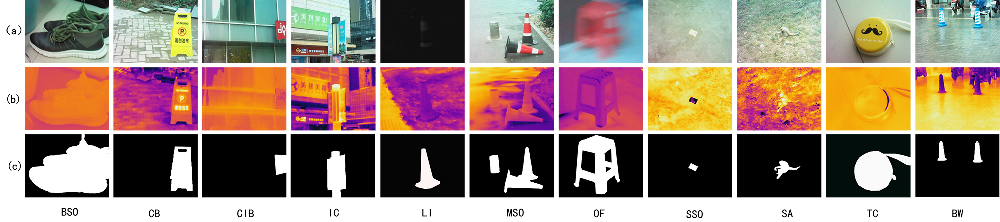}
	\caption{Sample image pairs with annotated ground truth and challenges from our RGBT dataset. (a) and (b) indicate the RGB image and its corresponding thermal image respectively. (c) is the corresponding ground truth of RGBT image pair.}
	\label{challenge_data}
\end{figure*}
\subsection{Dataset Statistics}
The image pairs in our dataset are recorded in different places and environments, moreover, our dataset records different illuminations, categories, sizes, positions and quantities of objects, as well as the backgrounds, etc. In specific, the following main aspects are considered when creating VT5000 dataset.

\textbf{Size of Object:} We define the size of the salient object as the ratio of number of pixels in the salient object to sum of all pixels in the image. If this ratio is more than 0.26, the object belongs to the big salient object.

\textbf{Illumination Conditions:} We create the image pairs under different light conditions(e.g., low-illumination, sunny or cloudy). Low-illumination and illumination variation under different environments usually bring great challenges to visible light images.

\textbf{Center Bias:} Previous studies on visual saliency show that the center bias has been identified as one of the most significant biases in the saliency datasets~\cite{LiHKRY14}, which involves a phenomenon that people pay more attentions to the center of the screen~\cite{tatler2005visual}. As described in~\cite{HuangZ17}, the degree of center bias cannot be described by simply overlapping all the maps in the dataset.

\textbf{Amounts of Salient Object:} It is called multiple salient objects that  the salient objects in an image are more than one. We find that the images have less salient objects in existing RGBT SOD datasets. In VT5000 dataset, we capture 3 to 6 salient objects in an image for the challenge of multiple salient objects.

\textbf{Background factor:} We take two factors related to background into consideration. Firstly, it is a big challenge that the temperature or appearance of the background is similar to the salient object. Secondly, it is difficult to separate salient objects accurately from cluttered background.

Considering above-mentioned factors, together with the challenges in existing RGBT SOD datasets~\cite{Wang0MZTL18,abs-1905-06741}, we annotate 11 challenges for testing different algorithms, including big salient object (BSO), small salient object (SSO), multiple salient object (MSO), low illumination (LI), center bias (CB), cross image boundary (CIB), similar appearance (SA), thermal crossover (TC), image clutter (IC), out of focus (OF) and bad weather (BW).
Descriptions for these challenges are as follows:
\textbf{BSO:} size of the object is the ratio of number of pixels in the salient object to sum of all pixels in the image. If the ratio is more than 0.26, the object belongs to the big salient object.
\textbf{SSO:} size of the object is the ratio of number of pixels in the salient object to sum of all pixels in the image. If the ratio is less than 0.05, the object belongs to the small salient object.
\textbf{LI:} the images are collected in cloudy days or at night.
\textbf{MSO:} there are more than one salient object in an image.
\textbf{CB:} the salient object is far away from the center of the image.
\textbf{CIB:} the salient object crosses the boundaries of the image, therefore the image always contains part of the object.
\textbf{SA:} the salient object has a similar color to the background surroundings.
\textbf{TC:} the salient object has a similar temperature to other objects or its surroundings.
\textbf{IC:} the scene around the object is complex or the background is cluttered.
\textbf{OF:} the object in the image is out-of-focus, and the whole image is blurred.
\textbf{BW:} the images collected in rainy days or greasy weather.
In addition, we also label those images with good or bad imaging quality of objects in RGB modality (RGB) or Thermal modality (T) in the dataset for researches in the future.
\textbf{RGB:} the objects are not clear in RGB modality.
\textbf{T:} the objects are not clear in Thermal infrared modality.
We also show the attribute distributions on the VT5000 dataset as shown in Table~\ref{attributes}. 
We use this 2D array to present the number of samples with one or two attributes, as many samples have multiple attributes(challenges). For example, the number '371' in the first row and second column represents that there are 371 samples with big salient object(BSO) and center bias(CB) challenges simultaneously. There are also a few samples with more than two kinds of challenges, we do not further show them here.

\begin{figure}[htbp]
	\includegraphics[width=3.2in]{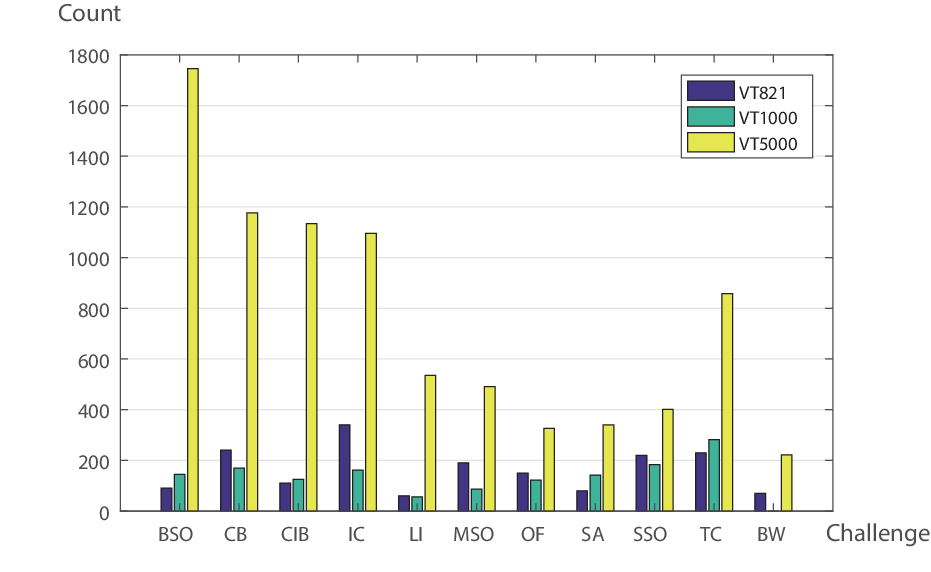}
	\caption{Challenge distribution of VT821, VT1000 and VT5000.}
	\label{motivation}
\end{figure}

The comparisons of our VT5000 with VT821 and VT1000 on the challenge distribution are shown in Fig.~\ref{motivation}. In addition, some challenging RGB and thermal infrared images in our dataset and the corresponding ground truths are shown in Fig.~\ref{challenge_data}.
\begin{figure}[htbp]
	\centering
	\includegraphics[width=3.2in]{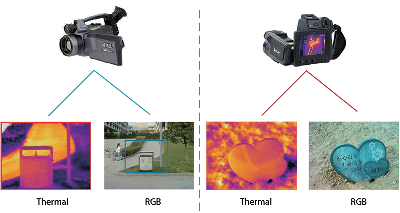}
	\caption{The image on the left shows a sample of RGBT image pairs from VT1000 dataset captured by FLIR (Forward Looking Infrared) SC620, the one on the right is a sample from VT5000 dataset captured by FLIR T640 and T610.}
	\label{camera}
\end{figure}

\begin{figure}[htbp]
	\centering
	\includegraphics[width=3.2in]{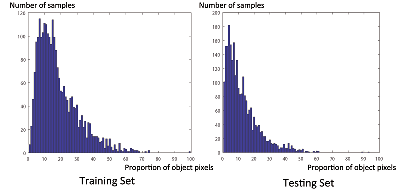}
	\caption{Size Distribution: the horizontal axis represents the proportion of pixels of objects to total pixels of the image, the vertical axis represents the number of corresponding images.}
	\label{size}
\end{figure}

Here, we also give another statistic result that is size distribution in Fig.~\ref{size}. It shows the distribution of size of the salient object in the training set and the testing set respectively. We can see that the big salient objects in the training set are more than those in the testing set, meanwhile the small salient objects in the testing set are more than those in the training set, which means our traing set is more generic while our testing contains more hard samples.

\subsection{Advantages of Our Dataset}
Generally speaking, it is possible to train a usable SOD model with 1000 or so samples. Therefore, our VT5000 with 2500 samples in training set and in testing set can be considered as a large scale dataset. 
Although existing VT821 and VT1000 are sufficient in quantity, their qualities are not satisfactory. The VT821 has many defective samples which catch unnatural noise and black padding caused by manual aligning. Image pairs in VT1000 are captured by a new equipment so that VT1000 avoids the influence of aligning. However, VT1000 contains too many simple samples. 
Our main purpose of constructing VT5000 is to boost the research for deep learning based RGBT SOD methods since existing VT821 and VT1000 are not suitable and sufficient for training a robust deep model and doing experiment analysis.
Different from VT821 and VT1000, we carefully set 11 challenges and collect samples in different environments. In addition, we further annotate the quality of two modalities, which may be useful for weakly supervised methods in future researches.
We divide all the data into training set and testing set for unifying experiment settings, which can effectively avoid unfair comparative experiments and reduce repetitive experiments for subsequent researches.
In sum, our VT5000 is more normative and valuable for studying and analyzing RGBT SOD methods.

Compared with existing RGBT datasets VT821 and VT1000, our VT5000 dataset has the following advantages: (1) Being different from previous thermal infrared camera as shown in Fig.~\ref{camera}, RGBT image pairs in our dataset do not require manual alignment, thus errors brought by manual alignment can be reduced; (2) Our thermal infrared camera can automatically focus, which enhances the accuracy of long-distance shooting and captures image texture information effectively; (3) Since these images were captured in summer and autumn, so our dataset has more thermal infrared images with severe thermal crossover; (4) We provide a large scale dataset with more RGBT image pairs and more complex scenes.

\begin{figure*}[htbp]
	\centering
	\includegraphics[width=7.0in]{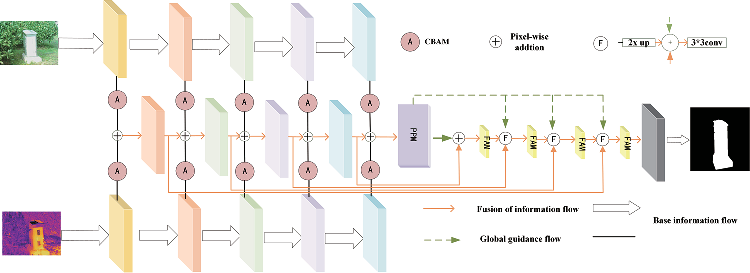}
	\caption{The overall architecture of our method. VGG16 is our backbone network, in which different color blocks represent different convolution blocks of VGG16.}
	\label{model}
\end{figure*}

\section{Attention-based Deep Fusion Network}
In this section, we will introduce the architecture of the proposed Attention-based Deep Fusion Network (ADFNet), and describe the details of RGBT salient object detection.

\subsection{Overview of ADFNet}
We build our network architecture based on ~\cite{simonyan2014very} as shown in Fig.~\ref{model}, and employ a two-stream CNN architecture, which first extracts RGB and thermal infrared features separately and then proceeds RGBT salient object detection.
We use the pre-trained VGG16 model as initialization, which provides a great ability for feature representation.
To make the network focus on more informative regions, we utilize a series of attention modules to extract weighted features from RGB and thermal infrared branches before fusion of these features.
From the second block of VGG16, the fused features of each layer are transmitted from the lower-level to the high-level in turn.
Although high-level semantic information could facilitate the location of salient  objects~\cite{WangZWL0RB18, LiuH16, HouCHBTT19}, low-level and mid-level features are also essential to refine deep level features.
Therefore, we add two complementary modules (Pyramid Pooling Module and Feature Aggregation Module)~\cite{abs-1904-09569} to accurately capture the exact position of a prominent object while sharpening its details.

\subsection{Convolutional Block Attention Module}
\begin{figure}[htbp]
	\centering
	\includegraphics[width=3.2in]{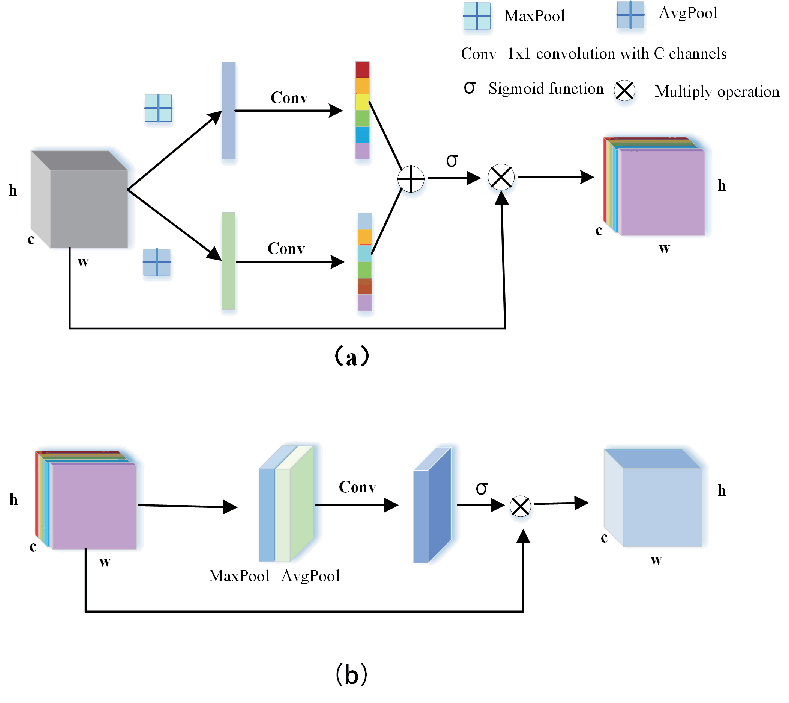}
	\caption{Convolutional Block Attention Module(CBAM), (a) channel-wise attention module and (b) spatial-wise attention module}
	\label{CBAM}
\end{figure}

\begin{figure}[htbp]
	\centering
	\includegraphics[width=3.2in]{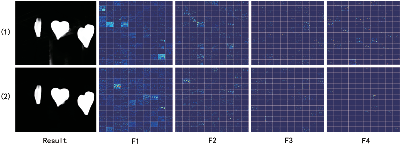}
	\caption{Visualization of features from different fusion layers in the proposed network without CBAM (shown in the first row) and with CBAM (shown in the second row). From left to right, there are the saliency map, the fused feature from layer 1 to 4, respectively.}
	\label{featuremaps}
\end{figure}

As illustrated in Fig.~\ref{model}, RGB and thermal infrared images respectively generate five different levels of features through five blocks of the backbone network VGG16, expressed by $X^{R}_{i}$ and $X^{T}_{i}$ $\in R^{C\times{H}\times{W}}$ respectively, where $i$ represents VGG16 $i$-th block.
As most of complex scenes contains cluttered background, which will bring lots of noises to feature extraction, we expect to selectively extract the features with less noises from RGB and thermal infrared branches.
Therefore, we adopt Convolutional Block Attention Module (CBAM)~\cite{WooPLK18} with channel-wise attention and spatial-wise attention shown in Fig.~\ref{CBAM}.
As shown in Fig.~\ref{featuremaps}, with CBAM, the proposed network can capture the spatial details around the objects, especially at the shallow layer, which is conducive to saliency refinement.
If without CBAM, the network will have some redundant information that is helpless for saliency refinement.

The channel-wise attention focuses on what makes sense for an input image.
Currently, most of methods typically uses average pooling operations to aggregate spatial information.
In addition to previous works~\cite{ZhaoW19,abs-1907-07449}, we think that the max-pooling collects discriminative characteristics of the object to infer finer channel-wise attention.
Therefore, we use both of average pooling and max pooling features.
The RGB branch is described here as an example, just as the thermal infrared branch does.
Firstly, we aggregate the spatial information from a feature map with the average pooling and max pooling operations to generate two different spatial context information, $X^{R_{avg}}_{i}$ and $X^{R_{max}}_{i}$, which represents the features after average pooling and max pooling respectively.
Secondly, these features are forwarded to two convolution layers of $1 * 1$ to generate channel attention map $M^{C_R}_{i}$, and we merge the outputted feature vectors with pixel-wise summation.
Finally, the channel attention weight vector is obtained by a sigmoid function.
The specific process can be expressed as:
\begin{align}
\label{channel_attention}
M^{C_R}_{i}=&(\sigma(Conv(AvgPool(X^{R}_{i})) \nonumber\\
&+Conv(MaxPool(X^{R}_{i}))))*X^{R}_{i}
\end{align}
where $\sigma$ denotes the sigmoid function, Conv denotes the convolutional operation and $*$ denotes multiply operation.

The spatial-wise attention is complementary to the channel attention.
Different from channel-wise attention, spatial-wise attention focuses on structural information and it highlights the informative spatial positions in the features.
In specific, we first apply average pooling and max pooling operations to features along the channel axis and connect these features to produce efficient descriptors.
Next, we obtain a two-dimensional feature map with a standard convolution layer. Finally, the spatial attention weight vector is obtained by a sigmoid function. The specific process can be expressed as:
\begin{equation}
\label{spatial_attention}
M^{S_R}_{i}=(\sigma({f^{k*k}([AvgPool(M^{C}_{i}),MaxPool(M^{C}_{i})])}))*M^{C_R}_{i}
\end{equation}
where $\sigma$ denotes the sigmoid function and $f^{k*k}$ represents a convolution operation with the filter size of $k*k$.


\subsection{Multi-modal Multi-layer Feature Fusion}
The previous works~\cite{abs-1806-01054,QiLJFU17} show that fusion of multi-modal features at the shallower layer or the deeper layer might not take good advantage of the useful features from multiple modalities.
To obtain rich and useful features of RGB and thermal infrared images during downsampling, we adopt a strategy of multi-layer feature fusion here.
Specifically, we use two VGG16 networks to extract RGB and thermal infrared features respectively, which can preserve RGB and thermal infrared features before upsampling.
Each branch provides a set of feature maps from each block of VGG16.
After passing through each Conv block, the two features are processed by CBAM respectively and then added for fusion on the pixel level.
Here, we add the features of two modalities directly in the first layer, and add the features of the current layer after the convolution operation with the features of the previous layer.
In this way, both of low-level features and high-level features are extracted, and the corresponding formula is expressed as:
\begin{equation}
\label{merged}
F_{i} =
\begin{cases}
M^{S_R}_{i}+M^{S_T}_{i}, & \mbox{if }i\mbox{ is 1} \\
Conv(F_{i-1})+M^{S_R}_{i}+M^{S_T}_{i}, & \mbox{if }i\mbox{ is 2,3,4,5}
\end{cases}
\end{equation}

\subsection{Pyramid Pooling Module}
A classic encoder-decoder classification architecture generally follows the top-down pathway. However, the top-down pathway is built upon the bottom-up backbone.
Higher-level features will be gradually diluted when they are transferred to shallower layers, therefore, loss of useful information inevitably happens.
The receptive field of CNN will become smaller and smaller when the number of network layers increases~\cite{ZhaoSQWJ17}, so the receptive field of the whole network is not large enough to capture the global information of the image.
Considering fine-level feature map lacks of high-level semantic information, we use a Pyramid Pooling Module (PPM)~\cite{abs-1904-09569} to process features for capturing global information with different sampling rates. Thus we can clearly identify position of the object at each stage.

More specifically, the PPM includes four sub-branches to capture context information of the image.
The first and fourth branches are the global average pooling layer and the identity mapping layer, respectively.
For the two intermediate branches, we use adaptive averaging pooling to ensure that sizes of output feature maps are 3*3 and 5*5 respectively.
The guidance information generated by PPM will be properly integrated with feature maps of different levels in the top-down pathway, and high-level semantic information can be easily passed to the feature map of each level by a series of up-sampling operations.
Providing global information to the feature of each level makes sure locating salient objects accurately.

\subsection{Feature Aggregation Module}
As shown in Fig.~\ref{model}, with the help of global guidance flow, the global guidance information can be passed to the feature at different pyramid level.
Next, we want to perfectly integrate the coarse feature map with the feature at different scale by the global guidance flow.
At first, the input image passes through five convolution blocks of VGG16 in sequence, thus feature maps corresponding to ${F = \{F_{2}, F_{3}, F_{4}, F_{5}\}}$ in the pyramid have been downsampled with downsample rate of $\{2, 4, 8, 16\}$ respectively.
In the original top-down pathway, RGB and thermal infrared features with coarser resolution are upsampled by a factor of 2.
After the merging operation, we use a convolutional layer with kernel size 3$\times$3 to reduce the aliasing effect of upsampling.

Here, we adopt a series of feature aggregation modules~\cite{abs-1904-09569}, each feature aggregation module contains four branches.
In the process of forward pass, with different downsampling rates, the input feature maps are first converted to the features with different scales by feeding them into an average pooling layer.
Then we combine the features from different branches through upsampling, followed by a 3$\times$3 convolutional layer, which helps our model reduce aliasing effects caused by upsampling operations, especially when the upsampling rate is large.

\subsection{Loss Function}
\subsubsection{Cross Entropy Loss}
The cross entropy loss is usually used to measure the error between the final saliency map and the ground truth of salient object detection. The cross entropy loss function is defined as:
\begin{equation}
\label{Cross_Entropy_Loss}
L_{C}=\sum^{size(Y)}_{i=0}({Y_i}log(P_i)+(1-Y_i)log(1-P_i))
\end{equation}
Where $Y$ represents the ground truth, $P$ represents the saliency map output by the network and $N$ represents the number of pixels in an image.

\begin{table*}[!ht]\normalsize
	\caption{List of the baseline methods with the main techniques and the published information}\label{baseline}
	\begin{center}
		\setlength{\abovecaptionskip}{35pt}
		\setlength{\belowcaptionskip}{35pt}
		\begin{tabular}{|c|c|c|c|}
			\hline
			{Baseline} &{Technique} &{Book Title } &{Year}\\
			\hline
			{RAS~\cite{ChenTWH18}} &residual learning and  reverse attention   &ECCV &2018\\
			{PiCANet~\cite{LiuH018}} &pixel-wise contextual attention network  &CVPR &2018\\
			{R3Net~\cite{deng2018r3net}} &recurrent residual refinement network &IJCAI  &2018\\
			{MTMR~\cite{Wang0MZTL18}} &multi task  manifold ranking  with cross-modality consistency  &IGTA &2018\\
			{SGDL~\cite{abs-1905-06741}} &collaborative graph learning algorithm  &TMM &2019\\
			{PFA~\cite{ZhaoW19}} &context-aware pyramid feature extraction module & CVPR &2019\\
			{CPD~\cite{WuSH19}} &multi-level feature aggregate &CVPR&2019\\
			{PoolNet~\cite{abs-1904-09569}} &global guidance module and feature aggregation module &CVPR &2019\\
			{BASNet~\cite{QinZHGDJ19}} &predict-refine architecture and a hybrid loss & CVPR &2019\\
			{EGNet~\cite{abs-1908-08297}} & integrate the local edge information and global location information &ICCV &2019\\
			\hline
		\end{tabular}
	\end{center}
\end{table*}
\begin{table*}[!ht]
	\caption{The value of F-measure in each challenge of our method and ten comparison methods}\label{challeng}
	\begin{center}
		\setlength{\abovecaptionskip}{5pt}
		\setlength{\belowcaptionskip}{5pt}
		\begin{tabular}{c|ccccccccccc}
			\hline
			{Challenge} &{PoolNet} &{BASNet} &{CPD} &{PFA}&{R3Net}&{RAS}&{PiCANet}&{EGNet}&{MTMR}&{SCGL}&{ADFNet}\\
			\hline
			{BSO} &0.800  &0.858 &0.872&0.802  &0.831  &0.768 &0.804 &0.873 &0.667 &0.754&\textbf{0.880}\\
			{CB} &0.725  &0.808 &0.845 &0.748 &0.794 &0.669 &0.796 &0.838 &0.575 &0.703&\textbf{0.854}\\
			{CIB} &0.740 &0.822  &0.860 &0.742 &0.822 &0.688 &0.790 &0.854 &0.582 &0.694&\textbf{0.860}\\
			{IC} &0.721  &0.775 &0.812 &0.735 &0.745 &0.672 &0.752 &0.818 &0.564 &0.681&\textbf{0.835}\\
			{LI} &0.757 &0.832   &0.840 &0.749 &0.790 &0.707 &0.783 &0.848 &0.695 &0.742&\textbf{0.868}\\
			{MSO} &0.706 &0.794   &0.826 &0.729 &0.774 &0.655 &0.777 &0.815 &0.620 &0.710&\textbf{0.837}\\
			{OF} &0.762 &0.816   &0.821 &0.754 &0.759 &0.738 &0.758 &0.817 &0.707 &0.738&\textbf{0.837}\\
			{SA} &0.727 &0.762   &0.825 &0.726 &0.728 &0.673 &0.748 &0.791 &0.653 &0.665&\textbf{0.835}\\
			{SSO} &0.658 &0.718   &0.767 &0.695 &0.663 &0.535 &0.676 &0.701 &0.698 &0.753&\textbf{0.806}\\
			{TC} &0.720 &0.791   &0.811 &0.762 &0.729 &0.711 &0.745 &0.791 &0.570 &0.675&\textbf{0.841}\\
			{BW} &0.750 &0.768   &0.795 &0.671 &0.753 &0.701 &0.773 &0.774 &0.606 &0.643&\textbf{0.804}\\
			{RGB} &0.733 &0.785   &0.804 &0.731 &0.736 &0.690 &0.743 &0.785 &0.670 &0.671&\textbf{0.817}\\
			{T} &0.719 &0.787   &0.802 &0.755 &0.719 &0.699 &0.736 &0.776 &0.564 &0.664&\textbf{0.833}\\
			\hline
		\end{tabular}
	\end{center}
\end{table*}

\subsubsection{Edge Loss}
The cross-entropy loss function provides a general guidance for the generation of the saliency map. Nevertheless, edge blur is an unsolved problem for saliency detection. Inspired by~\cite{ZhaoW19} and different from it, we use a simpler strategy to sharpen the boundary around the object. In specific, we use Laplace operator~\cite{gilbarg2015elliptic} to generate boundaries of ground truth and the predicted saliency map, and then we use the cross entry loss to supervise the generation of boundaries of salient object.
\begin{equation}
\label{gradient}
\Delta{f}=\frac{\partial^{2}f }{\partial{x}^2}+\frac{\partial^{2}f }{\partial{y}^2}
\end{equation}
\begin{equation}
\label{laplace}
\Delta\tilde{{f}}=abs(tanh(conv(f,K_{laplace})))
\end{equation}
\begin{equation}
\label{Edge_Loss}
L_{E}=-\sum^{size(Y)}_{i=0}(\Delta{Y_i}log(\Delta{P_i})+(1-\Delta{Y_i})log(1-\Delta{P_i}))
\end{equation}

The Laplace operator is defined as the divergence of the gradient $\Delta{f}$. Since the second derivative can be used to detect edges, we use the Laplace operator to obtain boundaries of salient object. Laplacian uses the gradient of image, which is actually calculated with convolution.  In Eq.~\ref{gradient}, $x$ and $y$ are the standard Cartesian coordinates of the $XY$-plane. Next, we use the absolute value operation followed by tanh activation in Eq.~\ref{laplace} to map the value to a range of 0 to 1. Then, we use the edge loss(Eq.~\ref{Edge_Loss}) to measure the error between real boundaries of salient object and its generated boundaries. The total loss can be represented as:
\begin{equation}
\label{total_Loss}
L_{S}=L_{C}+L_{E}
\end{equation}

\section{Experiments}
In this section, we first introduce our experiment setups, including the experimental details, the training set, the testing set, and the evaluation criteria. Then we conduct a series of ablation studies to prove the effect of each component in the proposed benchmark method. Finally, we show the performance of our method and compare it with the state-of-the-art methods.

To provide a comparison platform, Table~\ref{baseline} presents the baseline methods about the main technique, book title and published time. We take RGB and thermal images as the input to these ten state-of-the-art methods to achieve RGBT salient object detection, including PoolNet~\cite{abs-1904-09569}, RAS~\cite{ChenTWH18}, BASNet~\cite{QinZHGDJ19}, CPD~\cite{WuSH19}, R3Net~\cite{deng2018r3net}, PFA~\cite{ZhaoW19}, PiCANet~\cite{LiuH018}, EGNet~\cite{abs-1908-08297}, MTMR~\cite{Wang0MZTL18} and SGDL~\cite{abs-1905-06741}. These methods utilize deep features except for MTMR~\cite{Wang0MZTL18}. Furthermore, only MTMR~\cite{Wang0MZTL18} and SGDL~\cite{abs-1905-06741} are traditional models. In our method, we combine the deep features extracted from RGB and thermal branches and compare with the above-mentioned methods.

\subsection{Experiment Setup}
\textbf{Implementation Details.}
In this work, the proposed network is implemented based on the PyTorch and hyper-parameters are set as follows.
We train our network on single Titan Xp GPU.
We use Adam~\cite{KingmaB14} with a weight decay of 5e-4 to optimize parameters and train 25 epochs.
The initial learning rate is 1e-4, after the 20th epoch, the learning rate is reduced to 1e-5.
For the preprocessing of samples, we do data augmentation with simple random horizontal flipping.
The original size of input image is 640*480. To improve the efficiency of training stage, we resize the input image to 400*400.

In our code, we randomly initialize the parameters of network except of VGG16 backbone. 
We also set fixed random seeds for all random operations. So theoretically, the initialisation setups of each run are consistent. 
Although the results of different runs are not identical since the error of floating-point calculation, the difference of these results are very slight. 
So in subsequent experiments, we report the results of single run, which is representative and reasonable.

\textbf{Evaluation Metrics.} Similar to RGB dataset MSRA-B~\cite{liu2010learning}, we use the 2500 pairs of RGBT images in VT5000 dataset as the training set, and take the rest in VT5000 together with VT821~\cite{Wang0MZTL18} and VT1000~\cite{abs-1905-06741} as the test set.
We evaluate performances of different methods on three different metrics, including Precision-Recall (PR) curves, F-measure and Mean Absolute Error (MAE).
The PR curve is a standard metric to evaluate saliency performance, which is obtained by binarizing the saliency map using thresholds from 0 to 255 and then comparing the binary maps with the ground truth.
The F-measure can evaluate the quality of the saliency map, by computing the weighted harmonic mean of the precision and recall,
\begin{equation}
	\label{F-measure}
	F_{\beta}=\frac{(1+\beta^2)\cdot{Precision}\cdot{Recall}}{\beta^2\cdot{Precision}+{Recall}}
\end{equation}
where $\beta^2$ is set to 0.3 as suggested in~\cite{AchantaHES09}.
MAE is a complement to the PR curve and quantitatively measures the average difference between predicted $S$ and ground truth $G$ at the pixel level,
\begin{equation}
	\label{MAE}
	MAE=\frac{1}{W\times{H}}\sum^W_{x=1}\sum^H_{y=1}\mid{S(x,y)}-G(x,y)\mid
\end{equation}
where W and H is the width and height of a given image.

\begin{figure*}[htbp]
	\centering
	\includegraphics[width=7.0in]{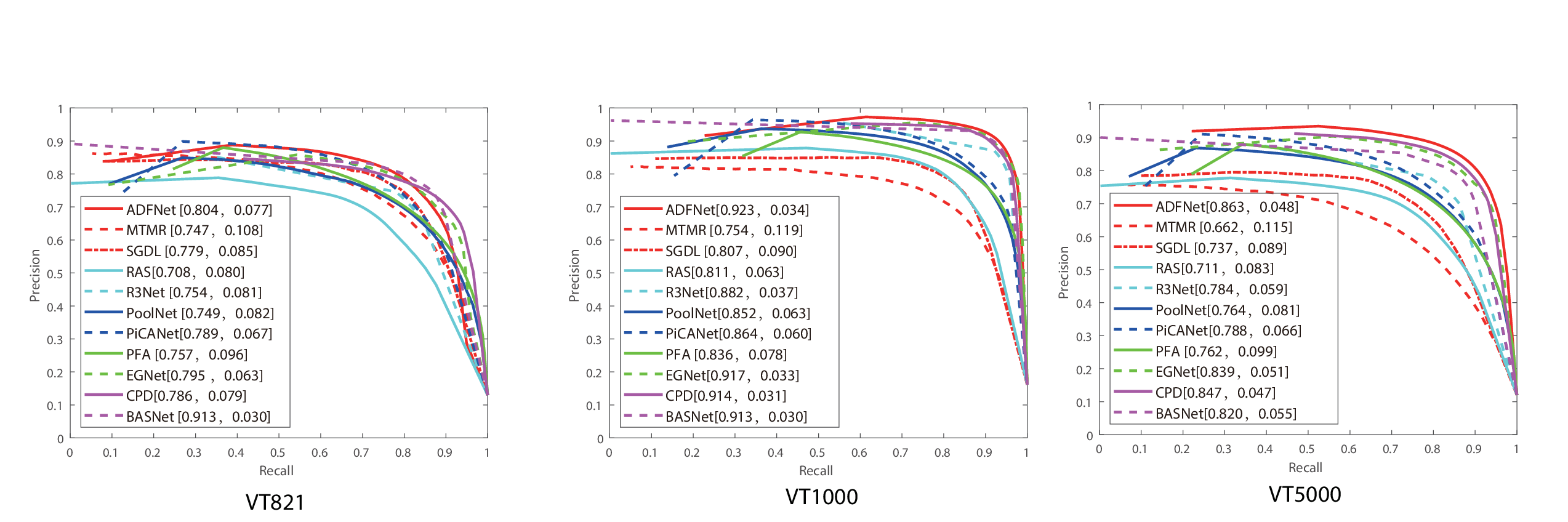}
	\caption{Precision-recall curves of our model compared with PoolNet~\cite{abs-1904-09569}, RAS~\cite{ChenTWH18}, BASNet~\cite{QinZHGDJ19}, CPD~\cite{WuSH19}, R3Net~\cite{deng2018r3net}, PFA~\cite{ZhaoW19}, PiCANet~\cite{LiuH018}, EGNet~\cite{abs-1908-08297}, MTMR~\cite{Wang0MZTL18}, SCGL~\cite{abs-1905-06741}. Our model can deliver state-of-the-art performance on three datasets. The numbers in first column are the values of F-measure, and the second column represents the value of MAE.}
	\label{PRF}
\end{figure*}

\subsection{Comparison with State-of-the-Art Methods}
We include eight deep learning-based and two traditional state-of-the-art methods in our benchmark for advanced evaluations, including PoolNet~\cite{abs-1904-09569}, RAS~\cite{ChenTWH18}, BASNet~\cite{QinZHGDJ19}, CPD~\cite{WuSH19}, R3Net~\cite{deng2018r3net}, PFA~\cite{ZhaoW19}, PiCANet~\cite{LiuH018}, EGNet~\cite{abs-1908-08297}, MTMR~\cite{Wang0MZTL18}, SCGL~\cite{abs-1905-06741}.
It is worth mentioning that results are obtained by testing the corresponding method on RGBT data without any post-processing, and evaluated with the same evaluation code. The results of all methods are obtained with the published codes.
For a fair comparison, the deep learning methods take the same training set and testing set as ours. 

\textbf{Challenge-sensitive performance.}
To dispaly and analyze the performance of our method on the challenge-sensitivity and imaging quality of objects compared with other methods, we give a quantitative comparison in Table~\ref{challeng}.
We evaluate our method on eleven challenges and bad imaging quality for objects in two modality (i.e., BSO, SSO, MSO, LI, CB, CIB, SA, TC, IC, OF, BW, RGB, T) in VT5000 dataset.
Notice that our method is significantly better than other methods, showing that our method is more robust for these challenges. Compared with PoolNet~\cite{abs-1904-09569}, our method outperforms 10.8\% and 14.8\% in F-measure on SA and SSO challengs, respectively. This results show that the thermal infrared data can provide effective information to help the network distinguish the object and the background when the object is similar to the background in RGB modality. And the small object is a challenge for every modality. Our network can locate salient object well with the help of the global guidance flow derived from PPM, even for the small object.

\textbf{Quantitative Comparisons.}
We compare the proposed methods with others in terms of F-measure scores, MAE scores, and PR-curves.
And we have verified the effectiveness of our method on three datasets, and the quantitative results are shown in Fig.~\ref{PRF}.
We take PoolNet~\cite{abs-1904-09569} as our baseline.
Fig.~\ref{PRF} shows the results on VT821, VT1000 and VT5000, and our method performs best in F-measure.
Compared with the baseline PoolNet, with the synergy of thermal infrared branch, our model outperforms PoolNet by a large margin of 5.5\%-9.9\% on three RGBT datasets(VT821, VT1000 and VT5000).
Compared with the method PiCANet~\cite{LiuH018} that also uses attention mechanism, our F-measure value achieves $1.5\%$ gains and MAE value is $0.1\%$ more than it on VT821 dataset.
On VT1000 dataset, our F-measure value outperforms
PiCANet~\cite{LiuH018} with $5.9\%$ and MAE value is $2.6\%$ less than it. 
On VT5000 dataset, our F-measure value outperforms PiCANet~\cite{LiuH018} with $7.5\%$ and MAE value is $1.8\%$ less than it.  
As a method with high performance, CPD~\cite{WuSH19} proposes a new Cascaded Partial Decoder framework for salient object detection, through integrating features of deeper layers and discarding larger resolution features of shallower layers to achieve fast and accurate salient object detection. Our F-measure value outperforms CPD~\cite{WuSH19} with $1.8\%$ and MAE value is $0.2\%$ less than it on VT821 dataset, our F-measure value outperforms it with $0.9\%$ and MAE value is $0.3\%$ less than it on VT1000 dataset. On VT5000 dataset, our F-measure value outperforms CPD~\cite{WuSH19} with $1.6\%$ and MAE value is $0.1\%$ more than it.
The EGNet~\cite{abs-1908-08297} is the latest approach among the compared methods, composed of three parts: edge feature extraction, feature extraction of salient object and one-to-one guidance module. The edge feature can help to locate the object and make the object boundary more accurate.
From Fig.~\ref{PRF}, we can see the results on VT1000 dataset.
And F-measure value of our method is $0.6\%$ higher than EGNet and its MAE value is $0.1\%$ lower than our method.
Same as above, the results of our method and other state-of-the-arts are shown in Fig.~\ref{PRF}(right). Compared with the baseline PoolNet, our F-measure value achieves $9.9\%$ gains and MAE value is $3.3\%$ less, and compared with best method EGNet~\cite{abs-1908-08297} , F-measure value of our method is $2.4\%$ higher than EGNet~\cite{abs-1908-08297} and its MAE value is $0.3\%$ lower than our method on VT5000. Compared with EGNet~\cite{abs-1908-08297}, our method has these merits: (1) Our method can refine the edge of the salient object without using additional edge detection model; (2) The global guidance flow derived from PPM can make good use of the global context information and better locate the salient object; (3) With the help of the thermal infrared branch, we can make use of the complementary information of the two modalities to better deal with various challenges in salient object detection.
This shows that our method is still optimal in general for RGBT SOD.

\textbf{PR Curves.} In addition to showing the results of F-measure and MAE, we also show the PR curves on three datasets. As shown in Fig.~\ref{PRF}, it can be seen that the PR curve (red) obtained by our method is particularly prominent compared with all previous methods. When the recall score is close to 1, our accuracy score is much higher than compared methods. This also shows that the truth-positive rate of our saliency maps is higher than compared methods.
\par
Our main purpose of constructing this dataset is to boost the researches on RGBT SOD since existing datasets are limited in scale so that they are not enough to train a robust deep learning model. 
Although we try our best to collect hard samples, there are still some normal samples. So the reported results do not show huge differences compared to VT821 and VT1000. 
From Fig.~\ref{PRF}, we can observe that the difficulty of VT5000 is at middle level among three datasets where the VT821 is hardest and the VT1000 is easiest, which is reasonable because the VT1000 collects an amount of simple samples while the samples in VT821 are randomly added noise and their quality is not well. Our VT5000 contains lots of challenging samples without any degradation operation.

\begin{figure*}[htbp]
	\centering
	\includegraphics[width=6.8in]{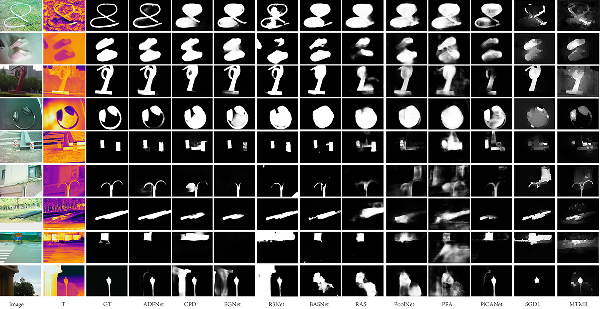}
	\caption{Saliency maps produced by the PoolNet~\cite{abs-1904-09569}, RAS~\cite{ChenTWH18}, BASNet~\cite{QinZHGDJ19}, CPD~\cite{WuSH19}, R3Net~\cite{deng2018r3net}, PFA~\cite{ZhaoW19}, PiCANet~\cite{LiuH018}, EGNet~\cite{abs-1908-08297}, MTMR~\cite{Wang0MZTL18}, SGDL~\cite{abs-1905-06741}. Our model can deliver state-of-the-art performance on three datasets}
	\label{saliency_map}
\end{figure*}

\textbf{Visual Comparison.} To qualitatively evaluate the proposed method on the new RGBT dataset, we visualize and compare some results of our method with other state-of-the-art methods in Fig.~\ref{saliency_map}. These examples are from various scenarios, including big salient object(BSO) (row 1, 3, 4, 7), multiple salient objects(MSO) (row 2, 5, 13), small salient object(SSO) (row 13), cross image boundary(CIB) (row 3, 6, 9),  cluttered background(IC)(row 3, 7, 9), low illumination(LI) (row 12), center bias(CB) (row 8), out-of-focus(OF) (row 2, 13), bad weather(BW) (row 3, 9), similar appearance(SA) (row 8) and Thermal Crossover(TC) (row 11, 13). Each row includes at least one challenge in Fig.~\ref{saliency_map}. It is easy to see that our method obtains best results in various challenging scenes. Specifically, the proposed method not only clearly highlights the objects, but also suppresses the background, and the objects have well-defined contours.

\section{Ablation Analysis}
In this part of ablation analysis, we respectively investigate the effectiveness of CBAM and edge loss of the proposed method. 
The experiments are performed on VT5000. We use the training set of VT5000 to train the model. Then we predict saliency maps of all the samples in testing set of VT5000 and compute the max F-measure and MAE for reporting.
As shown in Table~\ref{ablation}, firstly, we run the basic network without CBAM and edge loss and the result is not well. Then if we only add CBAM into the basic network PoolNet~\cite{abs-1904-09569}, the value of F-measure increases by 2.1\% and the value of MAE decreases by 0.5\%, and if we only add the edge loss, our performance is degraded. In the course of experiment, we find that if we only add the edge loss,  the value of loss is downward overall, but fluctuates greatly during training stage. 
In addition, although adding CBAM to the basic network can effectively suppress the noise, but if too much redundant noises appear, extracted edges are unsatisfactory, influencing greatly on the stability during training stage. 
These observations mean that without the attention mechanism, salient object can not be located accurately. 
As shown in Table~\ref{ablation}, with the help of CBAM for locating the salient object, and edge loss for refining edges, our network obtains best performances.

\begin{table}[!ht]
\caption{The impact of each component of the network on the performance}\label{ablation}
	\begin{center}
		\setlength{\abovecaptionskip}{5pt}
		\setlength{\belowcaptionskip}{5pt}
		\begin{tabular}{c|c|c|c}
			\hline
			{CBAM} &{Edge Loss} &{max $F_{\beta}$ } &{MAE}\\
			\hline
			{} &  &0.836 &0.057\\
			{\checkmark} & &0.857  &0.052\\
			{} &\checkmark  &0.825 &0.063\\
			{\checkmark} &\checkmark & \bfseries0.863 &\bfseries0.049\\
			\hline
		\end{tabular}
	\end{center}
\end{table}

We further study the time cost of all the modules(CBAM,PPM and FAM) used in our proposed model. We also do an ablation study on three modules and record corresponding training time and testing time.
The experiments are performed on our VT5000 dataset. And we report the training time of one epoch where 2500 samples are included in calculation.
The testing time is the time for computing VT5000's test set which also has 2500 samples.
As reported in Table~\ref{time cost}, the training time of CBAM,PPM and FAM separately increase about 0.095, 0.079 and 0.051 second per sample. And their testing time separately increase about 0.061, 0.042 and 0.025 second per sample.
Generally speaking, these modules will not bring too much time cost while boosting the performance.
\begin{table}[!ht]
\caption{The time cost of each module used in the network}\label{time cost}
	\begin{center}
		\setlength{\abovecaptionskip}{5pt}
		\setlength{\belowcaptionskip}{5pt}
		\begin{tabular}{c c c|c|c}
			\hline
			{CBAM} &{PPM} &{FAM} &{Training time} &{Testing time}\\
			\hline
			{} &  {\checkmark} &{\checkmark} &744.63s & 229.01s\\
			{\checkmark} &{}  &{\checkmark} &786.32s & 277.78s\\
			{\checkmark} &\checkmark  &{} &854.72s &321.32s\\
			{\checkmark} &{\checkmark}& {\checkmark}&982.65s &383.30s \\
			\hline
		\end{tabular}
	\end{center}
\end{table}

\section{Extension Experiments on Multi-modal SOD}
In this section, we further make analysis on multi-modal SOD that mainly includes RGBT SOD and RGBD SOD.
On one hand, we first apply existing RGBD SOD methods to our VT5000 dataset. Then we report max F-measure and MAE to show the performance of those methods.
On the other hand, we use our ADFNet to perform RGBD SOD task on existing RGBD datasets.

\subsection{The performance of RGBD SOD methods on RGBT datasets}
In recent years, RGBD SOD has been widely studied and many robust algorithms have been proposed.
The depth information is introduced as an assistant for boosting salient object detection in challenging scenes.
And RGBD SOD algorithm receives the inputs of two modalities, which is formally same to RGBT SOD task.
Therefore, to verify the effectiveness of our proposed method, we also need to compare with the modal-modal methods.
We select five advanced RGBD SOD methods which are publicly available, and train them on our VT5000 testing set with their default settings.
These five methods are DMRA\cite{piao2019depth}, MMCI\cite{chen2019multi}, AFNet\cite{wang2019adaptive}, A2dele\cite{piao2020a2dele} and S2MA\cite{liu2020learning}.

The performance comparison of these RGBD methods are shown in Table\ref{inRGBT}.  The advanced RGBD methods can also work well on RGBT data and our ADFNet achieves optimal or suboptimal results.
We don't make any specifical design for thermal image, and just propose a standard architecture with attention mechanism for multi-modal information aggregation.
The performances of some RGBD methods such as DMRA and A2dele are inferior to average level, which design a module for depth map or use the distillation strategy.
\begin{table}[!ht]
\renewcommand{\arraystretch}{1.1}
\setlength{\tabcolsep}{2.0mm}
\caption{The performance of RGBD methods on RGBT SOD datasets}
\label{inRGBT}
\begin{tabular}{c|cc|cc|cc}
\hline
\multirow{2}{*}{Methods} & \multicolumn{2}{c|}{VT821} & \multicolumn{2}{c|}{VT1000} & \multicolumn{2}{c}{VT5000}      \\ \cline{2-7} 
                         &{max $F_{\beta}$ } &{MAE}         &{max $F_{\beta}$ } &{MAE}     &{max $F_{\beta}$ } &{MAE}            \\ \hline
DMRA                     & 0.702        & 0.216       & 0.824             & 0.124   & 0.631          & 0.184          \\
MMCI                     & 0.723        & 0.089       & 0.875             & 0.040    & 0.797          & 0.056          \\
AFNet                    & 0.735        & 0.069       & 0.887             & 0.033   & 0.818          & 0.050          \\
A2dele                   & 0.644        & 0.074       & 0.813             & 0.061   & 0.725          & 0.065          \\
S2MA                     & 0.812        & 0.081       & 0.923             & 0.029   & 0.837          & 0.055          \\ \hline
ADF                      & 0.804        & 0.077       & 0.923    & 0.034  & 0.863   & 0.048 \\ \hline
\end{tabular}
\end{table}

\subsection{The performance of our method on RGBD SOD datasets}
For further exploring the effectiveness of our method on multi-modal task, we also conduct experiments on existing RGBD SOD datasets and compare our method with advanced RGBD SOD methods.
In this part, the compared methods are PCF\cite{chen2018progressively}, CTMF\cite{han2018cnns}, MMCI\cite{chen2019multi}, AFNet\cite{wang2019adaptive}, TANet\cite{chen2019three}, D3Net\cite{fan2020rethinking} and S2MA\cite{liu2020learning}.
The codes or saliency maps of these methods are publicly available.
We use DES\cite{cheng2014depth}, LFSD\cite{li2014saliency}, SSD\cite{zhu2017three} and STERE\cite{niu2012leveraging} as the test datasets. As same as the common setting of existing RGBD methods, we train our ADFNet on a data collection that contains randomly sampled 1485 image pairs and 700 image pairs from the NJU2K~\cite{ju2014depth} and NLPR~\cite{peng2014rgbd} datasets.
The introduction of above datasets are written in Section \ref{sec::2A}.

Compared with our ADFNet, the max F-measure and MAE scores of RGBD methods are reported in Table~\ref{inRGBD}. 
On three test datasets, our ADFNet shows similar performance to SOTA RGBD SOD methods, meaning that our methods can also be applied to other multi-modal SOD tasks with competitive results.
\begin{table}[ht]
\renewcommand{\arraystretch}{1.1}
\setlength{\tabcolsep}{0.7mm}
\caption{The quantitative comparison on RGBD SOD datasets}
\label{inRGBD}
\begin{tabular}{c|cccccccc}
\hline
\multirow{2}{*}{Methods}      & \multicolumn{2}{c|}{DES}            & \multicolumn{2}{c|}{LFSD}           & \multicolumn{2}{c|}{SSD}            & \multicolumn{2}{c}{STERE} \\ \cline{2-9} 
      & {max $F_{\beta}$ }    & \multicolumn{1}{c|}{MAE}   & {max $F_{\beta}$ }    & \multicolumn{1}{c|}{MAE}   & {max $F_{\beta}$ }    & \multicolumn{1}{c|}{MAE}   & {max $F_{\beta}$ }  & MAE   \\ \hline
PCF   & 0.842 & \multicolumn{1}{c|}{0.049} & 0.828 & \multicolumn{1}{c|}{0.112} & 0.843 & \multicolumn{1}{c|}{0.062} & 0.875       & 0.064       \\
CTMF  & 0.865 & \multicolumn{1}{c|}{0.055} & 0.814 & \multicolumn{1}{c|}{0.119} & 0.755 & \multicolumn{1}{c|}{0.099} & 0.848       & 0.086       \\
MMCI  & 0.839 & \multicolumn{1}{c|}{0.065} & 0.813 & \multicolumn{1}{c|}{0.132} & 0.823 & \multicolumn{1}{c|}{0.082} & 0.877       & 0.068       \\
AFNet & 0.775 & \multicolumn{1}{c|}{0.068} & 0.78  & \multicolumn{1}{c|}{0.133} & 0.735 & \multicolumn{1}{c|}{0.118} & 0.848       & 0.075       \\
TANet & 0.853 & \multicolumn{1}{c|}{0.046} & 0.827 & \multicolumn{1}{c|}{0.111} & 0.835 & \multicolumn{1}{c|}{0.063} & 0.878       & 0.060       \\
DMRA  & 0.906 & \multicolumn{1}{c|}{0.035} & 0.823 & \multicolumn{1}{c|}{0.111} & 0.874 & \multicolumn{1}{c|}{0.055} & 0.867       & 0.066       \\
S2MA  & 0.944 & \multicolumn{1}{c|}{0.021} & 0.862 & \multicolumn{1}{c|}{0.094} & 0.691 & \multicolumn{1}{c|}{0.138} & 0.895       & 0.051       \\ \hline
ADF   & 0.901   & \multicolumn{1}{c|}{0.038}   & 0.863   & \multicolumn{1}{c|}{0.098}   & 0.710   & \multicolumn{1}{c|}{0.120}   & 0.877         & 0.064         \\ \hline
\end{tabular}
\end{table}

\section{Concluding Remarks and Potential Directions }
In this work, we propose a novel attention-based deep fusion network for RGBT salient object detection.
Our network consists of a basic feature extraction network, convolutional block attention modules, pyramid pooling modules and feature aggregation modules.
The comparison experiments demonstrate our method performs best over all the state-of-the-art methods with most evaluation metrics.
We also create a new large-scale RGBT dataset for deep learning based RGBT salient object detection methods, with the attribute annotations for 11 challenges and the imaging quality annotations for salient objects in RGB and thermal images.
From the evaluation results,  taking advantage of the thermal image can boost the results of salient object detection in the scenarios of big salient objects, far away from center of the images, crossing the image boundaries, cluttered background, low illumination and similar temperature with background.
Cluttered background and low illumination are common scenes but bring big challenges to salient object detection, while thermal infrared images can provide complementary information to RGB images to improve SOD results.
However, when thermal crossover occurs, thermal data become unreliable, but visible spectrum imaging will not be influenced by temperature.

According to the evaluation results, we observe and draw some inspirations which are essential for boosting RGBT SOD in the future.
Firstly, deep learning-based RGBT SOD methods need to be explored further.
For example, how to design a suitable deep network which takes the special properties of RGB and thermal modalities into considerations for RGBT SOD is worth studying.
How to make best use of attention mechanisms and semantic information is still important for improving feature representations of salient objects and can prevent salient objects from being gradually diluted.
Secondly, the attribute-based feature representations could be studied for handling the problem of lacking sufficient training data.
Comparing with object detection and classification, the scale of annotated data for training networks of RGBT SOD are very small.
We annotate various attributes in our VT5000 dataset and these attribute annotations should be used to study the attribute-based feature representations, which can model different visual contents under certain attributes to reduce network parameters.
Thirdly, unsupervised and weakly supervised RGBT SOD are valuable research directions.
The task of RGBT SOD needs pixel-level annotations, thus annotating large-scale datasets needs unacceptable manual cost.
Therefore, no and less relying on large-scale labeled dataset is a potential research direction for RGBT SOD.
It should be noted that we have annotated some weakly supervised labels for our VT5000, i.e., imaging quality of different modalities, and we believe it would be beneficial to the related researches of unsupervised and weakly supervised RGBT SOD.
Finally, the alignment-free methods would make RGBT SOD more popular and practical in real-world applications.
We find that existing datasets contain some misaligned image pairs even though we adopt several advanced techniques to perform the alignment for images.
However, the images recorded from most of existing RGBT imaging platforms are non-aligned.
Therefore, alignment-free RGBT SOD is also worth being explored in the future.
\ifCLASSOPTIONcaptionsoff
  \newpage
\fi

\bibliographystyle{IEEEtran}
\bibliography{mybib}
\vspace{-15mm}
\begin{IEEEbiography}[{\includegraphics[width=1in,height=1.25in]{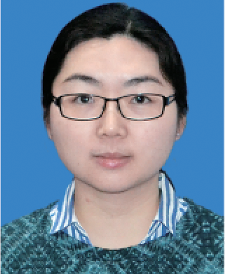}}]{Zhengzheng Tu}
received the M.S. and Ph.D.degrees from the School of Computer Science and Technology, Anhui University, Hefei, China, in
2007 and 2015, respectively. She is currently an Associate Professor with the School of Computer Science and Technology, Anhui University. Her current research interests include computer vision and deep learning.
\end{IEEEbiography}
\vspace{-15mm}
\begin{IEEEbiography}[{\includegraphics[width=1in,height=1.25in]{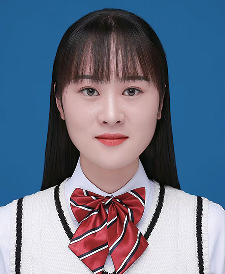}}]{Yan Ma}
received the M.S. degree from the School of Computer Science and Technology, Anhui University,
Hefei, China, in 2021. Her current research interests include computer vision and deep learning.
\end{IEEEbiography}
\vspace{-15mm}
\begin{IEEEbiography}[{\includegraphics[width=1in,height=1.25in]{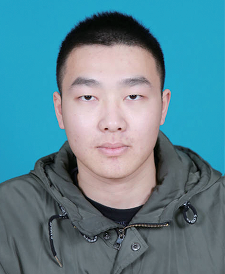}}]{Zhun Li}
received the B.Eng. degree in Anhui University, in 2019.
He is pursuing M.S. degree at the School of Computer Science and
Technology, in Anhui University, Hefei, China. His current
research interests include computer vision and deep learning.
\end{IEEEbiography}
\vspace{-15mm}
\begin{IEEEbiography}[{\includegraphics[width=1in,height=1.25in]{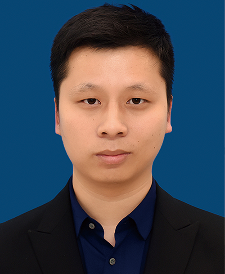}}]{Chenglong Li}
received the M.S. and Ph.D. degrees
from the School of Computer Science and Technology,
Anhui University, Hefei, China, in 2013 and
2016, respectively. From 2014 to 2015, he worked as
a Visiting Student with the School of Data and Computer
Science, Sun Yat-sen University, Guangzhou,
China. He was a postdoctoral research fellow at the
Center for Research on Intelligent Perception and
Computing (CRIPAC), National Laboratory of Pattern
Recognition (NLPR), Institute of Automation,
Chinese Academy of Sciences (CASIA), China. He
is currently an Associate Professor at the School of Computer Science and
Technology, Anhui University. His research interests include computer vision
and deep learning. He was a recipient of the ACM Hefei Doctoral Dissertation
Award in 2016.
\end{IEEEbiography}
\vspace{-15mm}
\begin{IEEEbiography}[{\includegraphics[width=1in,height=1.25in]{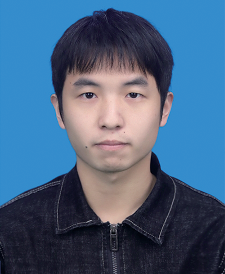}}]{Jieming Xu}
received the B.Eng. degree from Anhui University, Hefei, China, in 2021. He is currently pursuing the M.S. degree in College of Computer Science and Technology, Sichuan University, Chengdu, China. His current research interests include bigdata and deep learning.
\end{IEEEbiography}
\vspace{-15mm}
\begin{IEEEbiography}[{\includegraphics[width=1in,height=1.25in]{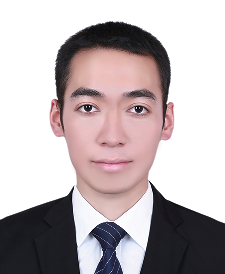}}]{Yongtao Liu}
received the B.Eng. degree from Anhui University, Hefei, China, in 2021.
His current research interests include computer vision and deep learning.
\end{IEEEbiography}

\end{document}